\documentclass[runningheads]{llncs}
\usepackage[T1]{fontenc}
\usepackage{graphicx}
\usepackage{amsmath}
\usepackage{float}
\usepackage{threeparttable}
\usepackage{makecell}
\usepackage{booktabs}
\usepackage{multirow}
\usepackage{array}
\usepackage{hyperref}
\usepackage{color}

\begin{document}
\title{Warmth and Competence in the Swarm: \\ Designing Effective Human-Robot Teams}
\titlerunning{Warmth and Competence in the Swarm}
%
\author{Genki Miyauchi\inst{1}\orcidID{0000-0002-3349-6765} \and
Roderich Gro\ss\inst{1,2}\orcidID{0000-0003-1826-1375} \and
Chaona Chen\inst{3}\orcidID{0000-0001-7289-1242}}
\index{Miyauchi, Genki}
\index{Gro\ss, Roderich}
\index{Chen, Chaona}
\authorrunning{G. Miyauchi et al.}
%
\institute{School of Electrical and Electronic Engineering, University of Sheffield, Sheffield, UK 
\email{g.miyauchi@sheffield.ac.uk} \and
Department of Computer Science, Technical University of Darmstadt, Darmstadt, Germany \email{roderich.gross@tu-darmstadt.de} \and 
School of Computer Science, University of Sheffield, Sheffield, UK
\email{chaona.chen@sheffield.ac.uk}}
 \maketitle              
\begin{abstract}
As groups of robots increasingly collaborate with humans, understanding how humans perceive them is critical for designing effective human-robot teams. While prior research examined how humans interpret and evaluate the abilities and intentions of individual agents, social perception of robot teams remains relatively underexplored. Drawing on the competence–warmth framework, we conducted two studies manipulating swarm behaviors in completing a collective search task and measured the social perception of swarm behaviors when human participants are either observers (Study 1) and operators (Study 2). Across both studies, our results show that variations in swarm behaviors consistently influenced participants' perceptions of warmth and competence. Notably, longer broadcast durations increased perceived warmth; larger separation distances increased perceived competence. Interestingly, individual robot speed had no effect on either of the perceptions. Furthermore, our results show that these social perceptions predicted participants’ team preferences more strongly than task performance. Participants preferred robot teams that were both warm and competent, not those that completed tasks most quickly. These findings demonstrate that human-robot interaction dynamically shapes social perception, underscoring the importance of integrating both technical and social considerations when designing robot swarms for effective human-robot collaboration.

\end{abstract}

\section{Introduction}

Swarm robotics is a rapidly growing field in which multiple robots work together as a team~\cite{dorigo2021swarm,dorigo2020reflections}. Unlike individual agents that maximize their own performance, robots in swarms prioritize collective goals, coordinating their actions to achieve outcomes that exceed the capability of any single robot~\cite{hamann2018swarm,cazenille2025signalling,karaguzel2023collective,sun2023mean}. As robots increasingly collaborate with humans, recent studies have explored human-swarm teaming, focusing primarily on improving objective performance metrics~\cite{kolling2016Human,miyauchi2023sharing,orozco2024extracting,heuthe2024counterfactual,ordaz2024improving,shan2024distributed}. Less attention has been given to how humans perceive and interact with swarms, despite the importance of such perceptions in shaping trust~\cite{nam2019models,orozco2024extracting} and preferences for specific agents~\cite{mckee2024warmth,dahiya2023survey,mckee2023humans,harris2023social}.

Humans naturally evaluate others along the dimensions of warmth and competence~\cite{fiske2007universal}. Competence reflects ability (e.g., intelligence, skill), while warmth reflects intent (e.g., friendliness, helpfulness). These dimensions also influence interactions with individual robots. For example, previous research suggest that agents perceived as both competent and warm are more trusted, whereas highly competent but low-warmth agents may be approached cautiously~\cite{mckee2023humans,harris2023social,oliveira2019stereotype,scheunemann2020warmth,li2022human}. Prior work has largely focused on single-agent systems~\cite{mckee2024warmth,scheunemann2020warmth}, but interactions with swarms may differ~\cite{dahiya2023survey,kaduk2024one}. Understanding whether warmth and competence extend to human-swarm interaction is critical for effective swarm design.

In this paper, we make the following contributions\footnote{A selection of preliminary findings were reported in~\cite{miyauchi2026human}.}:\mbox{}\par
\textbf{Social Perception Framework in Human-Swarm Interaction.} We investigate warmth and competence as key dimensions of social perception in robot swarm behaviors, and examine how they shape both human perceptions of and interactions with robot swarms.

\begin{figure}[t]
    \centering
    \includegraphics[width=\linewidth]
    {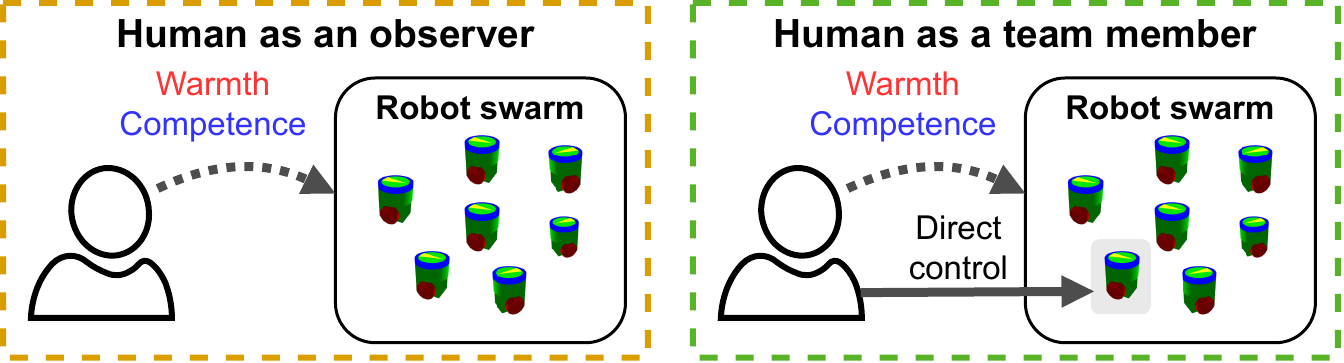}
    \caption{Two studies were conducted where participants rated the perceived warmth and competence of swarms of robots. In the first (left), they only observe the swarm. In the second (right), they also control a member of the swarm.}
    \label{fig:teaser-summary}
\end{figure}

\textbf{Influence of Human Involvement on Social Perception of Swarms.} Across Studies 1 and 2 (see Fig.~\ref{fig:teaser-summary}), we show that overall perceptions of robot swarms remain consistent across conditions, but human involvement as team members shifted their perception of the swarm behaviors. This highlights the dynamic nature of social perception in human-swarm interactions.

\textbf{Team Preferences Beyond Objective Metrics.} While both task performance and social perceptions influence team preference, we show that social perception has a stronger effect, highlighting the importance of designing robot swarms that prioritize socially preferred behaviors alongside task performance.

\textbf{Initial Guidance for Designing Socially Preferred Swarm Behaviors.} By manipulating robot parameters including speed, separation, and broadcast duration, we generated distinct swarm behaviors that were consistently perceived as highly warm and competent. These findings provide a foundation for the future design of socially preferred robot swarms.

\textbf{Swarm User Interface.} We built SwarmUI, a user interface supporting human-swarm collaboration and controlled studies of social perception by manipulating swarm behaviors. SwarmUI will be released as an open-source tool to facilitate future research on human-swarm interaction.

This paper is organized as follows. Section~\ref{sec:related-work} reviews related work on human social perception and human-swarm interaction. Section~\ref{sec:methods} introduces SwarmUI and the swarm behavior designs. Sections~\ref{sec:study1} and \ref{sec:study2} present Studies 1 and 2, respectively. Finally, Sections~\ref{sec:discussion} and \ref{sec:conclusion} conclude with discussion and implications.

\section{Related Work}
\label{sec:related-work}

\subsection{Fundamental Dimensions of Social Perception: Warmth and Competence} 

Research shows that warmth and competence are core dimensions of social perception shaping interactions at both individual and group levels~\cite{fiske2007universal,cuddy2008warmth,czopp2015positive}. These perceptions, sometimes expressed as stereotypes or biases~\cite{fiske2018model,cuddy2007bias,bargh1999cognitive,hentschel2019multiple} strongly influence engagement: people prefer interacting with those seen as both warm and competent, while competent but cold individuals may face cautious engagement or exclusion~\cite{fiske2007universal,fiske1999dis}. Notably, prior work shows that judgments of warmth and competence can depend on an individual's involvement in a group---for example, as an observer or as a team member. Individuals tend to perceive ingroup members as warmer than outgroup members, even when objective behavior is identical~\cite{cuddy2007bias,fiske2018model}. Research further indicates that judgments of competence and warmth are based on distinct factors. Competence is typically linked to an individual’s or group’s ability to perform tasks and tends to be relatively stable over time; a generally competent person or team may still be perceived as capable even after occasional failures. In contrast, warmth reflects moral and prosocial intentions, such as helping, cooperating, or prioritizing others’ needs---which can be evaluated more independently of task performance~\cite{fiske2007universal,fiske2018model}.

\subsection{Social Perception in Human-Swarm Interaction}

A key characteristic of swarm robotics is its inherent robustness as a group, demonstrated in three ways: the swarms are typically relatively homogeneous, improving their tolerance to failures in individual members; control is decentralized, removing a single point of failure; and the swarm can self-organize to adapt dynamically to changing situations~\cite{winfield2025ethical}. With the rapid advancement of swarm robotics, particularly in their ability to autonomously accomplish complex tasks, new challenges arise in designing systems that effectively support human-swarm collaboration. Recent research has primarily focused on enhancing the technical capabilities of swarm robots to enable collaborative tasks with single or multiple human partners~\cite{jang2021Omnipotent,miyauchi2023sharing}. Some studies in human-swarm interaction have focused on social perception through trust~\cite{abu2025towards}, psychophysiological effects~\cite{podevijn2016investigating}, and users' calibration of reliance on swarm performance and its impact on task outcomes~\cite{wilson2023Trustworthy,lyons2025examining}. Other studies have focused on the expressive qualities of swarm motion, showing that coordinated movement patterns can convey ``affective states'' such as happiness or sadness~\cite{dietz2017human,st2019collective,santos2021motions,kaduk2024emotional}. However, it remains unclear how humans perceive the broader social characteristics of robot swarms, when behaviors are systematically varied, and whether these perceptions differ between observers and active team members.

To address these questions, we developed SwarmUI, a swarm user interface and paired this with physics-based simulations in which a swarm of robots collaboratively search for and approach locations of interest. Building on prior work in swarm behavior design~\cite{kegeleirs2019random,miyauchi2023sharing}, we created distinct robot team behaviors and investigated for the first time social perceptions across two studies (see Fig.~\ref{fig:teaser-summary}): one in which participants acted as observers (Study 1) and one in which participants served as active team members (Study 2).

\section{Methods}
\label{sec:methods}

\subsection{Swarm Behavior Design}
\label{sec:swarm-design}

We simulated a homogeneous swarm of 10 robots using the ARGoS simulator~\cite{pinciroli2012argos}. We used this swarm size to ensure observable collective behavior while keeping the swarm visually manageable for participants. Each robot is based on the e-puck~\cite{mondada2009puck}, a mobile robot that has been widely used in swarm robotics due to its compact size, modular sensors, and reliable locomotion. The robot has a diameter of 7\,cm and moves using a differential-wheel drive. It is equipped with eight proximity sensors positioned along its perimeter. We assume the robot can sense its global position but cannot communicate globally. Instead, it is equipped with a range-and-bearing board that enables local communication with neighboring robots (up to a range of 36\,cm between the robot centers).

\begin{figure}
    \centering
    \includegraphics[width=\linewidth]{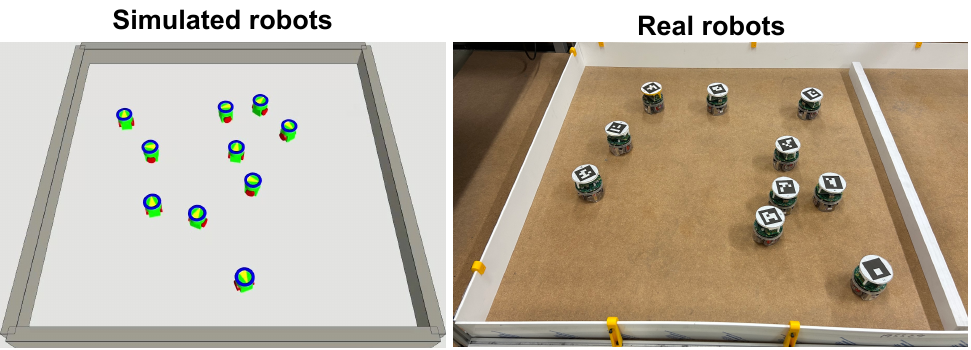}
    \caption{Overview of the experimental setup. Left: We investigate swarms of simulated robots that explore a bounded environment to discover a hidden target region (not shown). Right: Real-world equivalent of the setup.}
    
    \label{fig:setup}
\end{figure}

An overview of the experimental setup is shown in Fig.~\ref{fig:setup}. The robots reside in a square arena of side length 150\,cm, starting from uniformly random positions and orientations. Three seconds into the trial, a target region representing a location of interest appears at a random position. It is defined as a circular area of radius 25\,cm. 
Each robot detects the position of the target region if its body resides therein.
However, the robot is unaware of the locations of other robots. 
The objective for the swarm of robots is to all enter the target region as quickly as possible. To do so, each robot needs to explore the environment, discover the target either by itself or with the help from a team member, and move to it.

To enable all robots to efficiently arrive at the target location, we instructed the robots to perform the following three behaviors. The robots indicated their current behavior using distinct LED light patterns.

\textbf{Explore.} 
Each robot individually explores the arena using \textit{ballistic motion}~\cite{kegeleirs2019random}.
In this mode, a robot moves straight at a constant speed $v$ and rotates on the spot for a random duration when it encounters an obstacle, such as walls (i.e., the arena boundary) or other robots.
The direction of rotation is determined by which side the obstacle was detected; the robot will turn left if the obstacle was detected on the right side and vice versa.
The walls are detected using proximity sensors with a range of 10\,cm. Each robot maintains a desired separation distance $d$ from other robots, measured using its range-and-bearing sensor.
We used ballistic motion because it covers bounded areas more effectively than other random walk strategies while being simple to control~\cite{kegeleirs2019random}.

\textbf{Share target.}
Once a robot discovers the target region, it starts sharing the target's location via local communication with neighboring robots for a predefined duration of $T$ seconds.
During this duration, the robot continues to move exactly as it would in the \textit{Explore} mode.
Any robot that is informed for the first time of the location of the target will also start sharing the location for $T$ seconds.
This approach helps to spread the information across the swarm.
    
\textbf{Move to target.}
After sharing the target’s position for $T$ seconds, the robot moves toward the target location.
To do so, it realizes a flocking motion based on virtual potential forces, similarly to~\cite{miyauchi2022multi}, where robots are attracted to a common target center while being repelled by each other.
This allows the swarm of robots to cluster inside the target region regardless of the separation distance $d$, while avoiding collisions.

\begin{figure*}[t]
    \centering
    \includegraphics[width=\linewidth]{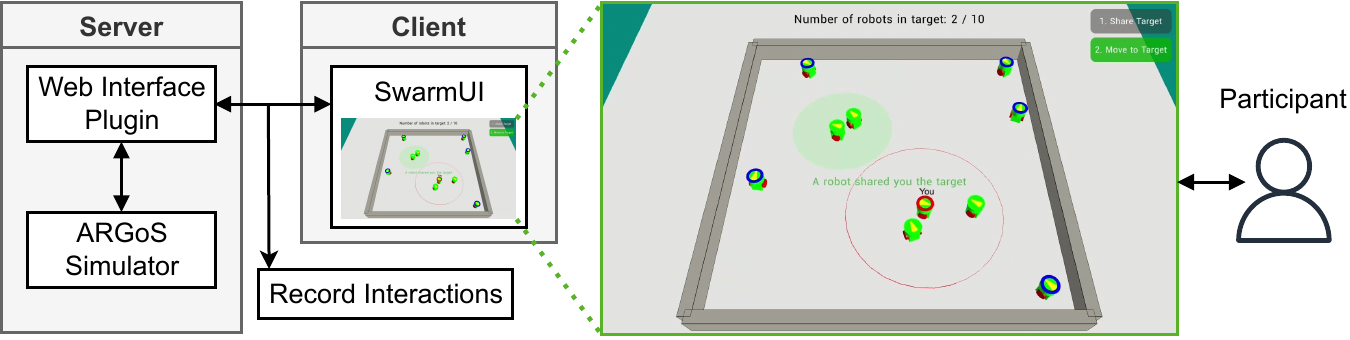}
    \caption{Overview of the system architecture, including user interface SwarmUI as used in Study 2. The participant interacts with the swarm simulation via the user interface. They control one robot of the swarm.}
    \label{fig:study2_system_overview}
\end{figure*}

To elicit a range of group behaviors in the robot swarm, we systematically varied three core parameters of the aforementioned exploration strategy: \textit{speed} $v\in\{5.0, 7.5, 10.0, 12.5, 15.0\}$\,cm/s, \textit{separation distance:} $d\in\{4, 12, 20, 28, 36\}$\,cm, and \textit{local broadcast duration:} $T\in\{0, 4, 8, 12, 16\}$\,s.
The parameters were selected based on the physical capabilities of the e-puck robots and their expected impact on swarm behavior. The speed range reflects the feasible locomotion performance of the e-puck in real-world settings, ensuring that simulated behavior can be transferred to physical robots. Separation distance captures variations in the robots’ ability to maintain spacing with each other, which directly influences group-level coordination. Local broadcast duration defines the temporal window during which a focal robot locally shares information with its peers, and varying it allows us to modulate how the information is spread across the swarm. Combining the five levels of each parameter yielded 125 robot team configurations. \textit{Task performance} was measured as the time required for all robots in each team to reach the target region, with shorter times indicating better performance. Parameter checks confirmed that the selected parameters and their ranges produced a wide distribution of completion times (13--103\,s, mean around 60\,s) and that each parameter significantly affected task performance.

\subsection{Swarm User Interface, SwarmUI}

\label{sec:user_interface}
To allow users to interact with the swarm as team members (Study 2), we developed SwarmUI, a custom user interface for human-swarm interaction (see Fig.~\ref{fig:study2_system_overview}).
Each participant controlled one robot via a web-based ARGoS plugin~\cite{patel2022multi}, while the remaining robots operated autonomously as described in Section~\ref{sec:swarm-design}. 
The source code is available from~\cite{miyauchi_gross_chen2025_source}.

As shown in the screenshot of SwarmUI (See Fig.~\ref{fig:study2_system_overview}), the user-controlled robot was labeled “You” and surrounded by a red circle indicating its communication range. By default, the user-controlled robot remained in \textit{Explore} mode, moving straight forward. Participants could steer the robot by holding the left or right keyboard keys, rotating it in place. This allowed users to control the robot’s heading while maintaining the same speed as the remaining robots of the swarm. Participants can also switch between the three behaviors described in Section~\ref{sec:swarm-design} using buttons at the top-right corner of SwarmUI. Pressing the button \textit{“Share Target”} enabled the robot to broadcast the target location once it had discovered it, either directly, or through local communication by a peer. Pressing \textit{“Move to Target”} directed the robot to navigate autonomously to and remain inside the target area for the remainder of the trial. Participants were instructed to press “Move to Target” only after they were satisfied that the target location had been shared with other robots and wanted their robot to move directly to the target.

\subsection{Experimental Design}

To examine the impact of human involvement on perceptions of robot swarms, we conducted two studies where participants acted as observers or active team members (see Fig.~\ref{fig:teaser-summary}). Although a within-subjects design could allow direct comparisons within individuals, we ran separate studies to (1) recruit a larger, more diverse sample in Study 1 to identify features perceived as warm or competent, and (2) avoid carryover effects between roles~\cite{ho2025causal,keren2014between}, ensuring clear separation and complementary insights across the studies.

\section{Study 1: Perception of Robot Swarm Behaviors as an Observer}
\label{sec:study1}

In Study 1, we aimed to examine whether the participants perceive different levels of warmth and competence when observing different robot swarm behaviors, and whether these social perceptions influence their preferences for selecting robots as potential teammates. Preliminary findings were reported in~\cite{miyauchi2026human}.

\subsection{Participants}

We recruited 90 online participants (hereafter referred to as observers) via Prolific (45 female, 45 male; $M_{\text{age}} = 29.6 \text{ yrs}$, $SD_{\text{age}} = 4.5 \text{ yrs}$). The sample size was chosen to ensure that each robot team was rated by at least 10 observers, providing sufficient data for stable estimation of stimulus-level effects in subsequent analysis. All observers reported normal or corrected-to-normal vision. Written informed consent was obtained from each observer prior to the study, and observers were compensated at a rate of £12/hour. The experimental protocol was approved by University of Sheffield Ethics Committee (Reference ID: 068772).

\subsection{Procedure}

Each observer first watched a short demonstration to familiarize themselves with the experimental setting and robot behaviors described in Section~\ref{sec:swarm-design} (see Supplementary Video for an example clip). They were informed that the goal was for all robots to have reached the target region as quickly as possible. Observers then watched robot teams performing the task and rated each team on perceived warmth, competence, and team preference. We assessed perceptions of warmth (“How friendly, approachable, or cooperative is the group of robots?”) and competence (“How capable, effective, and intelligent is the group of robots?”) using 7-point Likert scales. The scale descriptions were adapted from~\cite{mckee2024warmth}, which evaluated the same dimensions in human–agent interaction. Each observer rated 15 randomly selected teams, with trials presented in random order. To reduce order effects, the rating sequence (i.e., warmth, competence, team preference) was counterbalanced across participants. After rating all teams, observers completed a brief post-task questionnaire and provided free-text responses describing robot behaviors they perceived as warm or competent.

\subsection{Results}

We first examined how swarm parameters (speed, separation, and broadcast) influenced perceptions of warmth and competence using a linear mixed-effects model (LME) (see Table~\ref{tab:lme_social_perception_with_parameters}). Warmth was significantly predicted by broadcast duration ($p < 0.001$), with longer information sharing perceived as warmer. Competence was significantly influenced by separation distance ($p = 0.01$), with more dispersed robots perceived as more competent. No parameter interactions were significant (all $p > 0.10$), and robot speed had no significant effect on either dimension.

\begin{table}[t]
    \centering
    \begin{threeparttable}
        \setlength{\tabcolsep}{6pt} 
        \caption{LME models of robot swarm parameters on warmth \& competence. SE = standard error; t = t-statistic. Adapted from~\cite{miyauchi2026human}.}
        \label{tab:lme_social_perception_with_parameters}
        \begin{tabular}{llcccc}
        \toprule
        & \textbf{predictor} & \textbf{estimate} & \textbf{SE} & \textbf{t} & \textbf{p-value} \\
        \midrule
        \multirow{3}{*}{\makecell[l]{model for \\ warmth ratings}} 
        & speed        & -0.002 & 0.020 & -0.111 & 0.912 \\
        & separation   & 0.011 & 0.006 & 1.814 & 0.070 \\
        & \textbf{broadcast} & \textbf{0.043} & \textbf{0.012} & \textbf{3.565} & \textbf{0.0004} \\
        \midrule
        \multirow{3}{*}{\makecell[l]{model for \\ competence ratings}} 
        & speed        & 0.014 & 0.022 & 0.619 & 0.536 \\
        & \textbf{separation}   & \textbf{0.018} & \textbf{0.007} & \textbf{2.564} & \textbf{0.010} \\
        & broadcast    & -0.013 & 0.014 & -0.903 & 0.367 \\
        \bottomrule
        \end{tabular}
        \begin{tablenotes}
            \footnotesize
            \item Significant predictors ($p < .05$) are highlighted in bold.
        \end{tablenotes}
    \end{threeparttable}
\end{table}

Previous research has shown that warmth and competence is a better predictor for participants’ team preferences in human-agent collaboration than objective performance~\cite{mckee2024warmth,harris2023social}, suggesting that humans may prioritize perceived social traits over task outcomes in their team preferences. To test this, we modeled how team preference is predicted by warmth, competence, and task performance, using a similar LME analysis as above. Task performance was measured as the time to complete the task and thus reversed for interpretability (i.e., higher values = better performance). All predictors were standardized (z-scored) to allow comparison of effect sizes. As shown in Table~\ref{tab:lme_team_pref}, social perception ratings and objective task performance positively predicted team preference ($p < 0.001$ for all). Warmth ($\beta = 0.74$) and competence ($\beta = 0.92$) emerged as strong predictors, whereas task performance showed a significant but smaller effect ($\beta = 0.12$). This indicates that observers’ preferences were driven more strongly by the social perceptions of the robot teams than by their objective performance.  

\begin{table}[t]
    \centering
    \begin{threeparttable}
        \caption{LME model predicting team preference from social perceptions and task performance (standardized). SE = standard error; t = t-statistic. Adapted from~\cite{miyauchi2026human}.}
        \label{tab:lme_team_pref}
        \setlength{\tabcolsep}{10pt} 

        \begin{tabular}{lcccc}
        \toprule
        \textbf{predictor} & \textbf{estimate} & \textbf{SE} & \textbf{t} & \textbf{p-value} \\
        \midrule
        \textbf{warmth}             & \textbf{0.739} & \textbf{0.030} & \textbf{24.73} & \textbf{< 0.001} \\
        \textbf{competence}         & \textbf{0.924} & \textbf{0.031} & \textbf{29.70} & \textbf{< 0.001} \\
        \textbf{task performance}   & \textbf{0.121} & \textbf{0.024} & \textbf{5.02}  & \textbf{< 0.001} \\
        \bottomrule
        \end{tabular}
        
        \begin{tablenotes}
            \footnotesize
            \item Significant predictors ($p < .05$) are highlighted in bold.
        \end{tablenotes}
    \end{threeparttable}
\end{table}

Based on Study~1, we first divided all robot teams into four quadrants (1) high warmth–high competence, (2) low warmth–high competence, (3) high warmth–low competence, and (4) low warmth–low competence, using the median ratings of warmth and competence across all participants. Within each quadrant, we then ranked teams by the sum of their averaged warmth and competence scores and selected the top ten teams to ensure that Study~2 included robot teams with distinct behaviors.

\section{Study 2: Perception of Robot Swarm Behaviors as an Operator}
\label{sec:study2}

\subsection{Participants}

We recruited 16 participants (8 female, 8 male; $M_{\text{age}} = 26.8 \text{ yrs}$, $SD_{\text{age}} = 9.9 \text{ yrs}$). A priori power analysis for a repeated-measures ANOVA ($\alpha = 0.05$, power = 0.8) indicated that this sample size would be sufficient to detect medium-sized effects (i.e., $f = 0.25$). The inclusion criteria and ethical approval were identical to those described in Study 1.

\subsection{Procedure}

Each participant came to the lab and completed the task in person. The task instructions and the description of the social perception measures were kept consistent across all participants, following the same procedure as in Study~1. In each round, participants directly controlled one robot while the remaining robots were operated by the system using the same parameters as in Study 1. Participants used SwarmUI introduced in Section \ref{sec:user_interface}. They were instructed that the goal was for all robots to reach the target region as quickly as possible, with each round limited to 60 seconds. Following the procedure of a previous study on social perception in human–agent interactions~\cite{mckee2024warmth}, participants received a performance-based bonus of £0.50 for each successful round. In each round, participants rated the team on perceived warmth, competence, and joint effort. Here, we measured \textit{joint effort} instead of team preference, because participants interacted with a new team in every round (i.e., irrelevant to their preference). The measurement of joint effort captured their sense of belonging to the team. Each participant completed 20 rounds, comprising five robot teams randomly sampled from each of the four groups identified in Study~1.

\subsection{Results}

Following the experiments for Study~2, we first examined whether participants’ social perceptions of the teams were preserved overall. We included all trials in the analysis, except for one trial from a participant who did not successfully complete the task. Table \ref{tab:social_perception_study2} shows participants’ ratings of warmth and competence for each group, averaged across robot teams and participants (1 = very low, 7 = very high). A one-way ANOVA revealed significant differences in Study 2 ratings of both warmth and competence across the four groups (Warmth: $F(3, 316) = 10.93$, $p < 0.001$; Competence: $F(3, 316) = 19.99$, $p < 0.001$). Post-hoc Tukey HSD tests further confirmed significant differences in ratings for intended high vs. low warmth and competence teams, supporting the robustness of our results. These results suggest the robot swarm behaviors were robust across conditions, whether participants acted as observers or team members.

\begin{table}[t]
\centering
\caption{Averaged ratings of warmth and competence in human---robot teamwork.}
\label{tab:social_perception_study2}
\begin{threeparttable}
\setlength{\tabcolsep}{5pt} 
\begin{tabular}{>{\bfseries}l c c}
\toprule
Group & Warmth (M±SD) & Competence (M±SD) \\
\midrule
HighWarmth\_HighCompetence & 5.01 (1.50) & 5.91 (1.08) \\
LowWarmth\_HighCompetence  & 3.90 (1.82) & 5.09 (1.57) \\
HighWarmth\_LowCompetence  & 5.18 (1.37) & 4.30 (1.52) \\
LowWarmth\_LowCompetence   & 4.38 (1.66) & 4.59 (1.43) \\
\midrule
\textbf{ANOVA F (df)} & 10.93 (3,316) & 19.99 (3,316) \\
\textbf{p-value}       & < .001        & < .001        \\
\bottomrule
\end{tabular}
\end{threeparttable}
\end{table}

\begin{figure}[t]
    \centering
\includegraphics[width=\linewidth]
{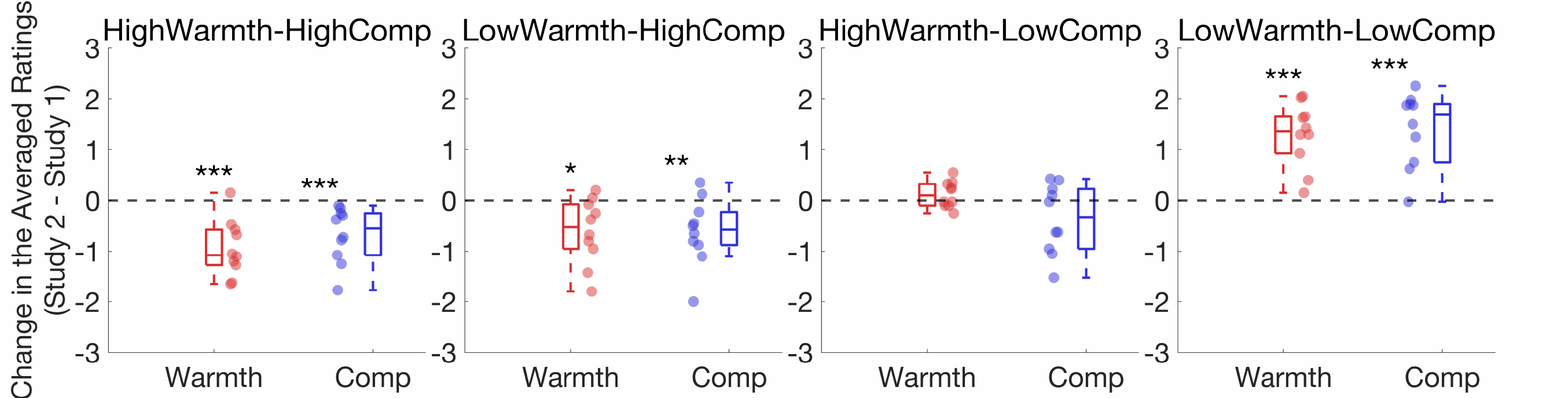}
    \caption{Boxplots show the changes in warmth (red) and competence (blue) ratings between Study 1 and Study 2, with each dot representing an individual robot swarm team. Significance levels: *$p<0.05$, **$p<0.01$, ***$p<0.001$.}
    \label{fig:study2_summary}
\end{figure}

Although overall perceptions of the swarms were similar to Study 1, participants’ perceptions shifted when they actively joined the teams. To assess this, we compared Study 2 ratings with Study 1 ratings for each group using two-sample t-tests. Fig.~\ref{fig:study2_summary} shows each group (labeled at the top of each subplot), with changes in warmth (red) and competence (blue) ratings. While general perceptions were largely preserved, interacting with the robots as team members shifted social perception. For instance, high-warmth, high-competence teams (leftmost subplot) were rated lower on both dimensions ($p < 0.001$), whereas low-warmth, low-competence teams (rightmost subplot) were rated higher ($p < 0.001$).

\section{Discussion}
\label{sec:discussion}

Across two studies, we investigated how participants perceive warmth and competence in robot swarms under observation and direct control. Although direct control slightly modulated social perceptions of swarm behavior, the overall pattern of high versus low warmth and competence remained consistent across conditions. These findings demonstrate the robustness of our parameterized swarm behaviors across human–robot interaction contexts.

\textit{Social Perceptions of Robot Swarm Behaviors.} Our results show that robot swarm behaviors elicit robust social perceptions, extending prior work on single agents~\cite{scheunemann2020warmth,mckee2024warmth,mckee2023humans,harris2023social,liu2022friendly}. While swarm-robotics research often prioritizes movement speed~\cite{dietz2017human,song2024speed,kaduk2024one}, our findings highlight two other parameters that significantly shape social perception. Within the parameter range tested in our study, larger separation distances were perceived as more competent, likely because the more distributed formations suggested improved exploration capability~\cite{bacula2022motis,levillain2018more}. Longer broadcasting durations increased perceived warmth, as sustained broadcasting was often interpreted as robots “helping” each other~\cite{mieczkowski2019helping,carpinella2017robotic} and signaled joint engagement~\cite{le2024sense,loehr2022sense,navare2024performing}. These results offer practical guidance for designing more socially aware human–swarm interactions. Qualitative analysis of the post-task questionnaire responses revealed that participants consistently associated perceived warmth with prosocial, collective behaviors. In particular, warmth judgments were linked to robots’ behaviors to share information and prioritize group success over individual efficiency. For example, participants highlighted behaviors associated with warmth: \textit{``Robots went to share the target with others''}, \textit{``I considered whether the robots which found the targets first would help other robots''}, and \textit{``not going to the target straight away and moving around to share''}.

\textit{Social Perception Over Objective Performance in Team Preference.} Our results show that social perceptions more strongly influenced team preferences than task performance. These results extend single-agent research to swarms, highlighting that optimizing task performance alone is insufficient for effective human-robot collaboration~\cite{mckee2024warmth,harris2023social,scheunemann2020warmth,pizzi2023chatbot}. Designers should consider behaviors that convey warmth and collaboration, informed by more fine-grained, multidimensional assessments of social perception~\cite{carpinella2017robotic,nomura2006measurement}. Future work can draw on human social behavior, natural systems, and variations in swarm size~\cite{podevijn2016investigating} to design swarms that are both effective and socially engaging~\cite{mitri2013using,sevillano2016warmth}.

\section{Conclusion}
\label{sec:conclusion}

In summary, our findings demonstrate that robot swarm behaviors elicit social perceptions of warmth and competence, which exert a stronger influence on human team preferences than task performance. Although active engagement in human–robot teams modulates these perceptions, robot parameters, including broadcasting duration and inter-robot distance, reliably shaped participants’ perceptions during observation and collaboration. These findings underscore the importance of integrating technical and social considerations when designing robot swarms to ensure they are not only effective but also socially preferred.

\begin{credits}
\subsubsection{\ackname} 
This research was supported by the Horizon Europe \href{https://openswarm.eu}{OpenSwarm} project (Grant No. 101093046) and \href{https://robotics-institute-germany.de/}{Robotics Institute Germany} (BMBF Grant No. 16ME1001). The views expressed do not necessarily reflect those of the funders.
\subsubsection{\discintname}
The authors have no competing interests to declare.
\end{credits}
%
%
\newpage
\bibliographystyle{splncs04}
\bibliography{references}

@article{pinciroli2012argos,
  title={{ARGoS}: A modular, parallel, multi-engine simulator for multi-robot systems},
  author={Pinciroli, Carlo and Trianni, Vito and O’Grady, Rehan and Pini, Giovanni and Brutschy, Arne and Brambilla, Manuele and Mathews, Nithin and Ferrante, Eliseo and Di Caro, Gianni and Ducatelle, Frederick and others},
  journal={Swarm intelligence},
  volume={6},
  number={4},
  pages={271--295},
  year={2012},
  publisher={Springer}
}

@inproceedings{mondada2009puck,
  title={The e-puck, a robot designed for education in engineering},
  author={Mondada, Francesco and Bonani, Michael and Raemy, Xavier and Pugh, James and Cianci, Christopher and Klaptocz, Adam and Magnenat, Stephane and Zufferey, Jean-Christophe and Floreano, Dario and Martinoli, Alcherio and others},
  booktitle={Proceedings of the 9th conference on autonomous robot systems and competitions},
  volume={1},
  pages={59--65},
  year={2009},
  organization={Castelo Branco: IPCB, Instituto Polit{\'e}cnico de Castelo Branco}
}

@inproceedings{kegeleirs2019random,
  title={Random walk exploration for swarm mapping},
  author={Kegeleirs, Miquel and Garz{\'o}n Ramos, David and Birattari, Mauro},
  booktitle={Annual conference towards autonomous robotic systems},
  pages={211--222},
  year={2019},
  organization={Springer}
}

@inproceedings{miyauchi2022multi,
  title = {Multi-Operator Control of Connectivity-Preserving Robot Swarms Using Supervisory Control Theory},
  author = {Miyauchi, Genki and Lopes, Yuri K and Gro{\ss}, Roderich},
  booktitle = {2022 International Conference on Robotics and Automation (ICRA)},
  pages = {6889--6895},
  year = {2022},
  organization = {IEEE},
}

@inproceedings{miyauchi2023sharing,
  title = {Sharing the Control of Robot Swarms Among Multiple Human Operators: A User Study},
  author = {Miyauchi, Genki and Lopes, Yuri K and Gro{\ss}, Roderich},
  booktitle = {2023 IEEE/RSJ International Conference on Intelligent Robots and Systems (IROS)},
  pages = {8847--8853},
  year = {2023},
  organization = {IEEE},
}

@article{patel2022multi,
  title={On multi-human multi-robot remote interaction: A study of transparency, inter-human communication, and information loss in remote interaction},
  author={Patel, Jayam and Sonar, Prajankya and Pinciroli, Carlo},
  journal={Swarm Intelligence},
  volume={16},
  number={2},
  pages={107--142},
  year={2022},
  publisher={Springer}
}

@article{nomura2006measurement,
  title={Measurement of negative attitudes toward robots},
  author={Nomura, Tatsuya and Suzuki, Tomohiro and Kanda, Takayuki and Kato, Kensuke},
  journal={Interaction Studies. Social Behaviour and Communication in Biological and Artificial Systems},
  volume={7},
  number={3},
  pages={437--454},
  year={2006},
  publisher={John Benjamins Publishing Company Amsterdam/Philadephia}
}

@article{kolling2016Human,
  title = {Human Interaction With Robot Swarms: A Survey},
  author = {Kolling, A and Walker, P and Chakraborty, N and Sycara, K and Lewis, M},
  year = {2016},
  volume = {46},
  pages = {9--26},
  publisher = {{IEEE}},
  issn = {VO - 46},
  journal = {IEEE Transactions on Human-Machine Systems},
  number = {1},
}

@article{nam2019models,
  title={Models of trust in human control of swarms with varied levels of autonomy},
  author={Nam, Changjoo and Walker, Phillip and Li, Huao and Lewis, Michael and Sycara, Katia},
  journal={IEEE Transactions on Human-Machine Systems},
  volume={50},
  number={3},
  pages={194--204},
  year={2019},
  publisher={IEEE}
}

@ARTICLE{dorigo2021swarm,
  author={Dorigo, Marco and Theraulaz, Guy and Trianni, Vito},
  journal={Proceedings of the IEEE}, 
  title={Swarm Robotics: Past, Present, and Future [Point of View]}, 
  year={2021},
  volume={109},
  number={7},
  pages={1152-1165},
}

@book{hamann2018swarm,
  title={Swarm robotics: A formal approach},
  author={Hamann, Heiko},
  volume={221},
  year={2018},
  publisher={Springer}
}

@article{santos2021motions,
  title={From motions to emotions: Can the fundamental emotions be expressed in a robot swarm?},
  author={Santos, Mar{\'\i}a and Egerstedt, Magnus},
  journal={International Journal of Social Robotics},
  volume={13},
  number={4},
  pages={751--764},
  year={2021},
  publisher={Springer}
}

@article{podevijn2016investigating,
  title={Investigating the effect of increasing robot group sizes on the human psychophysiological state in the context of human--swarm interaction},
  author={Podevijn, Ga{\"e}tan and O’grady, Rehan and Mathews, Nithin and Gilles, Audrey and Fantini-Hauwel, Carole and Dorigo, Marco},
  journal={Swarm Intelligence},
  volume={10},
  number={3},
  pages={193--210},
  year={2016},
  publisher={Springer}
}

@article{st2019collective,
  title={Collective expression: How robotic swarms convey information with group motion},
  author={St-Onge, David and Levillain, Florent and Zibetti, Elisabetta and Beltrame, Giovanni},
  journal={Paladyn, Journal of Behavioral Robotics},
  volume={10},
  number={1},
  pages={418--435},
  year={2019},
  publisher={Sciendo}
}

@inproceedings{kaduk2024emotional,
  title={Emotional tandem robots: How different robot behaviors affect human perception while controlling a mobile robot},
  author={Kaduk, Julian and Weilbeer, Friederike and Hamann, Heiko},
  booktitle={2024 IEEE/RSJ International Conference on Intelligent Robots and Systems (IROS)},
  pages={2465--2470},
  year={2024},
  organization={IEEE}
}

@article{lyons2025examining,
  title={Examining the human-centred challenges of human--swarm interaction},
  author={Lyons, Joseph B and Capiola, August and Adams, Julie A and Mator, Janine D and Cherry, Erin and Barrera, Kristen},
  journal={Philosophical Transactions A},
  volume={383},
  number={2289},
  pages={20240140},
  year={2025},
  publisher={The Royal Society},
}

@article{winfield2025ethical,
  title={On the ethical governance of swarm robotic systems in the real world},
  author={Winfield, Alan FT and Swana, Matimba and Ives, Jonathan and Hauert, Sabine},
  journal={Philosophical Transactions A},
  volume={383},
  number={2289},
  pages={20240142},
  year={2025},
  publisher={The Royal Society}
}

@inproceedings{dietz2017human,
  title={Human perception of swarm robot motion},
  author={Dietz, Griffin and E, Jane L and Washington, Peter and Kim, Lawrence H and Follmer, Sean},
  booktitle={Proceedings of the 2017 CHI conference extended abstracts on human factors in computing systems},
  pages={2520--2527},
  year={2017}
}

@inproceedings{wilson2023Trustworthy,
  title     = {Trustworthy Swarms},
  booktitle = {Proceedings of the {{First International Symposium}} on {{Trustworthy Autonomous Systems}}},
  author    = {Wilson, James and Chance, Greg and Winter, Peter and Lee, Suet and Milner, Emma and Abeywickrama, Dhaminda and Windsor, Shane and Downer, John and Eder, Kerstin and Ives, Jonathan and Hauert, Sabine},
  year      = {2023},
  month     = jul,
  series    = {{{TAS}} '23},
  pages     = {1--11},
  publisher = {{Association for Computing Machinery}},
}

@inproceedings{jang2021Omnipotent,
  title     = {Omnipotent Virtual Giant for Remote Human-Swarm Interaction},
  booktitle = {2021 30th {{IEEE International Conference}} on {{Robot}} \& {{Human Interactive Communication}} ({{RO-MAN}})},
  author    = {Jang, Inmo and Hu, Junyan and Arvin, Farshad and Carrasco, Joaquin and Lennox, Barry},
  year      = {2021},
  pages     = {488--494},
  issn      = {1944-9437},
}

@article{dahiya2023survey,
  title={A survey of multi-agent human--robot interaction systems},
  author={Dahiya, Abhinav and Aroyo, Alexander M and Dautenhahn, Kerstin and Smith, Stephen L},
  journal={Robotics and Autonomous Systems},
  volume={161},
  pages={104335},
  year={2023},
  publisher={Elsevier}
}

@article{mckee2024warmth,
  title={Warmth and competence in human-agent cooperation},
  author={McKee, Kevin R and Bai, Xuechunzi and Fiske, Susan T},
  journal={Autonomous Agents and Multi-Agent Systems},
  volume={38},
  number={1},
  pages={23},
  year={2024},
  publisher={Springer}
}

@article{dorigo2020reflections,
  title={Reflections on the future of swarm robotics},
  author={Dorigo, Marco and Theraulaz, Guy and Trianni, Vito},
  journal={Science robotics},
  volume={5},
  number={49},
  pages={eabe4385},
  year={2020},
  publisher={American Association for the Advancement of Science}
}

@article{sun2023mean,
  title={Mean-shift exploration in shape assembly of robot swarms},
  author={Sun, Guibin and Zhou, Rui and Ma, Zhao and Li, Yongqi and Gro{\ss}, Roderich and Chen, Zhang and Zhao, Shiyu},
  journal={Nature Communications},
  volume={14},
  number={1},
  pages={3476},
  year={2023},
  publisher={Nature Publishing Group UK London}
}

@article{orozco2024extracting,
  title={Extracting Human Levels of Trust in Human--Swarm Interaction Using {EEG} Signals},
  author={Orozco, Jesus A and Artemiadis, Panagiotis},
  journal={IEEE Transactions on Human-Machine Systems},
  volume={54},
  number={2},
  pages={182--191},
  year={2024},
  publisher={IEEE}
}

@article{cazenille2025signalling,
  title={Signalling and social learning in swarms of robots},
  author={Cazenille, Leo and Toquebiau, Maxime and Lobato-Dauzier, Nicolas and Loi, Alessia and Macabre, Loona and Aubert-Kato, Nathana{\"e}l and Genot, Anthony J and Bredeche, Nicolas},
  journal={Philosophical Transactions A},
  volume={383},
  number={2289},
  pages={20240148},
  year={2025},
  publisher={The Royal Society}
}

@article{karaguzel2023collective,
  title={Collective gradient perception with a flying robot swarm},
  author={Karag{\"u}zel, Tugay Alperen and Turgut, Ali Emre and Eiben, AE and Ferrante, Eliseo},
  journal={Swarm Intelligence},
  volume={17},
  number={1},
  pages={117--146},
  year={2023},
  publisher={Springer}
}

@article{heuthe2024counterfactual,
  title={Counterfactual rewards promote collective transport using individually controlled swarm microrobots},
  author={Heuthe, Veit-Lorenz and Panizon, Emanuele and Gu, Hongri and Bechinger, Clemens},
  journal={Science Robotics},
  volume={9},
  number={97},
  pages={eado5888},
  year={2024},
  publisher={American Association for the Advancement of Science}
}

@article{ordaz2024improving,
  title={Improving performance in swarm robots using multi-objective optimization},
  author={Ordaz-Rivas, Erick and Torres-Trevi{\~n}o, Luis},
  journal={Mathematics and Computers in Simulation},
  volume={223},
  pages={433--457},
  year={2024},
  publisher={Elsevier}
}

@article{shan2024distributed,
  title={A distributed multi-robot task allocation method for time-constrained dynamic collective transport},
  author={Shan, Xiaotao and Jin, Yichao and Jurt, Marius and Li, Peizheng},
  journal={Robotics and Autonomous Systems},
  volume={178},
  pages={104722},
  year={2024},
  publisher={Elsevier}
}

@article{mckee2023humans,
  title={Humans perceive warmth and competence in artificial intelligence},
  author={McKee, Kevin R and Bai, Xuechunzi and Fiske, Susan T},
  journal={Iscience},
  volume={26},
  number={8},
  year={2023},
  publisher={Elsevier}
}

@article{harris2023social,
  title={Social perception in Human-{AI} teams: Warmth and competence predict receptivity to AI teammates},
  author={Harris-Watson, Alexandra M and Larson, Lindsay E and Lauharatanahirun, Nina and DeChurch, Leslie A and Contractor, Noshir S},
  journal={Computers in Human Behavior},
  volume={145},
  pages={107765},
  year={2023},
  publisher={Elsevier}
}

@article{fiske2007universal,
  title={Universal dimensions of social cognition: Warmth and competence},
  author={Fiske, Susan T and Cuddy, Amy JC and Glick, Peter},
  journal={Trends in cognitive sciences},
  volume={11},
  number={2},
  pages={77--83},
  year={2007},
  publisher={Elsevier}
}

@inproceedings{oliveira2019stereotype,
  title={The stereotype content model applied to human-robot interactions in groups},
  author={Oliveira, Raquel and Arriaga, Patr{\'\i}cia and Correia, Filipa and Paiva, Ana},
  booktitle={2019 14th ACM/IEEE International Conference on Human-Robot Interaction (HRI)},
  pages={123--132},
  year={2019},
  organization={IEEE}
}

@inproceedings{scheunemann2020warmth,
  title={Warmth and competence to predict human preference of robot behavior in physical human-robot interaction},
  author={Scheunemann, Marcus M and Cuijpers, Raymond H and Salge, Christoph},
  booktitle={2020 29th IEEE international conference on robot and human interactive communication (RO-MAN)},
  pages={1340--1347},
  year={2020},
  organization={IEEE}
}

@article{li2022human,
  title={How human-like behavior of service robot affects social distance: A mediation model and cross-cultural comparison},
  author={Li, Linyao and Li, Yi and Song, Bo and Shi, Zhaomin and Wang, Chongli},
  journal={Behavioral Sciences},
  volume={12},
  number={7},
  pages={205},
  year={2022},
  publisher={MDPI}
}

@inproceedings{kaduk2024one,
  title={From one to many: How active robot swarm sizes influence human cognitive processes},
  author={Kaduk, Julian and Cavdan, M{\"u}ge and Drewing, Knut and Hamann, Heiko},
  booktitle={2024 33rd IEEE International Conference on Robot and Human Interactive Communication (ROMAN)},
  pages={1207--1212},
  year={2024},
  organization={IEEE}
}

@article{cuddy2008warmth,
  title={Warmth and competence as universal dimensions of social perception: The stereotype content model and the BIAS map},
  author={Cuddy, Amy JC and Fiske, Susan T and Glick, Peter},
  journal={Advances in experimental social psychology},
  volume={40},
  pages={61--149},
  year={2008},
  publisher={Elsevier}
}

@incollection{fiske2018model,
  title={A model of (often mixed) stereotype content: Competence and warmth respectively follow from perceived status and competition},
  author={Fiske, Susan T and Cuddy, Amy JC and Glick, Peter and Xu, Jun},
  booktitle={Social cognition},
  pages={162--214},
  year={2018},
  publisher={Routledge}
}

@article{cuddy2007bias,
  title={The BIAS map: behaviors from intergroup affect and stereotypes.},
  author={Cuddy, Amy JC and Fiske, Susan T and Glick, Peter},
  journal={Journal of personality and social psychology},
  volume={92},
  number={4},
  pages={631},
  year={2007},
  publisher={American Psychological Association}
}

@article{fiske1999dis,
  title={({Dis}) respecting versus (dis) liking: Status and interdependence predict ambivalent stereotypes of competence and warmth},
  author={Fiske, Susan T and Xu, Juan and Cuddy, Amy C and Glick, Peter},
  journal={Journal of social issues},
  volume={55},
  number={3},
  pages={473--489},
  year={1999},
  publisher={Wiley Online Library}
}

@article{bargh1999cognitive,
  title={The cognitive monster: The case against the controllability of automatic stereotype effects.},
  author={Bargh, John A},
  journal={Dual-process theories in social psychology},
  year={1999},
  publisher={The Guilford Press}
}

@article{czopp2015positive,
  title={Positive stereotypes are pervasive and powerful},
  author={Czopp, Alexander M and Kay, Aaron C and Cheryan, Sapna},
  journal={Perspectives on Psychological Science},
  volume={10},
  number={4},
  pages={451--463},
  year={2015},
  publisher={Sage Publications Sage CA: Los Angeles, CA}
}

@article{hentschel2019multiple,
  title={The multiple dimensions of gender stereotypes: A current look at men’s and women’s characterizations of others and themselves},
  author={Hentschel, Tanja and Heilman, Madeline E and Peus, Claudia V},
  journal={Frontiers in psychology},
  volume={10},
  pages={11},
  year={2019},
  publisher={Frontiers Media SA}
}

@article{liu2022friendly,
  title={Friendly or competent? {The} effects of perception of robot appearance and service context on usage intention},
  author={Liu, Xing Stella and Yi, Xiao Shannon and Wan, Lisa C},
  journal={Annals of Tourism Research},
  volume={92},
  pages={103324},
  year={2022},
  publisher={Elsevier}
}

@article{mitri2013using,
  title={Using robots to understand social behaviour},
  author={Mitri, Sara and Wischmann, Steffen and Floreano, Dario and Keller, Laurent},
  journal={Biological Reviews},
  volume={88},
  number={1},
  pages={31--39},
  year={2013},
  publisher={Wiley Online Library}
}

@article{sevillano2016warmth,
  title={Warmth and competence in animals},
  author={Sevillano, Ver{\'o}nica and Fiske, Susan T},
  journal={Journal of Applied Social Psychology},
  volume={46},
  number={5},
  pages={276--293},
  year={2016},
  publisher={Wiley Online Library}
}

@article{pizzi2023chatbot,
  title={I, chatbot! The impact of anthropomorphism and gaze direction on willingness to disclose personal information and behavioral intentions},
  author={Pizzi, Gabriele and Vannucci, Virginia and Mazzoli, Valentina and Donvito, Raffaele},
  journal={Psychology \& Marketing},
  volume={40},
  number={7},
  pages={1372--1387},
  year={2023},
  publisher={Wiley Online Library}
}

@article{ho2025causal,
  title={Causal Inference in Counterbalanced Within-Subjects Designs},
  author={Ho, Justin and Min, Jonathan},
  journal={arXiv preprint arXiv:2505.03937},
  year={2025}
}

@incollection{keren2014between,
  title={Between-or within-subjects design: A methodological dilemma},
  author={Keren, Gideon},
  booktitle={A handbook for data analysis in the behaviorial sciences},
  pages={257--272},
  year={2014},
  publisher={Psychology Press}
}

@inproceedings{mieczkowski2019helping,
  title={Helping not hurting: Applying the stereotype content model and BIAS map to social robotics},
  author={Mieczkowski, Hannah and Liu, Sunny Xun and Hancock, Jeffrey and Reeves, Byron},
  booktitle={2019 14th ACM/IEEE International Conference on Human-Robot Interaction (HRI)},
  pages={222--229},
  year={2019},
  organization={IEEE}
}

@inproceedings{carpinella2017robotic,
  title={The robotic social attributes scale ({RoSAS}) development and validation},
  author={Carpinella, Colleen M and Wyman, Alisa B and Perez, Michael A and Stroessner, Steven J},
  booktitle={Proceedings of the 2017 ACM/IEEE International Conference on human-robot interaction},
  pages={254--262},
  year={2017}
}

@article{le2024sense,
  title={Sense of agency in joint action: A critical review of we-agency},
  author={Le Besnerais, Alexis and Moore, James W and Berberian, Bruno and Grynszpan, Ouriel},
  journal={Frontiers in psychology},
  volume={15},
  pages={1331084},
  year={2024},
  publisher={Frontiers Media SA}
}

@article{loehr2022sense,
  title={The sense of agency in joint action: An integrative review},
  author={Loehr, Janeen D},
  journal={Psychonomic Bulletin \& Review},
  volume={29},
  number={4},
  pages={1089--1117},
  year={2022},
  publisher={Springer}
}

@article{navare2024performing,
  title={When performing actions with robots, attribution of intentionality affects the sense of joint agency},
  author={Navare, Uma Prashant and Ciardo, Francesca and Kompatsiari, Kyveli and De Tommaso, Davide and Wykowska, Agnieszka},
  journal={Science Robotics},
  volume={9},
  number={91},
  pages={eadj3665},
  year={2024},
  publisher={American Association for the Advancement of Science}
}

@article{song2024speed,
  title={Speed and density planning for a speed-constrained robot swarm through a virtual tube},
  author={Song, Wenqi and Gao, Yan and Quan, Quan},
  journal={IEEE Robotics and Automation Letters},
  year={2024},
  publisher={IEEE}
}

@article{bacula2022motis,
  title={{MoTiS} parameters for expressive multi-robot systems: Relative motion, timing, and spacing},
  author={Bacula, Alexandra and Knight, Heather},
  journal={International Journal of Social Robotics},
  volume={14},
  number={9},
  pages={1965--1993},
  year={2022},
  publisher={Springer}
}

@inproceedings{levillain2018more,
  title={More than the sum of its parts: Assessing the coherence and expressivity of a robotic swarm},
  author={Levillain, Florent and St-Onge, David and Zibetti, Elisabetta and Beltrame, Giovanni},
  booktitle={2018 27th IEEE International Symposium on Robot and Human Interactive Communication (RO-MAN)},
  pages={583--588},
  year={2018},
  organization={IEEE}
}

@misc{miyauchi_gross_chen2025_source,
  author={Miyauchi, Genki and Gro{\ss}, Roderich and Chen, Chaona}, 
  title={{SwarmUI} and robot controller source code.},
  year={2025},
  url={https://github.com/genkimiyauchi/swarm-perception},
}

@inproceedings{abu2025towards,
  title={Towards Understanding the Impact of Swarm Motion on Human Trust},
  author={Abu-Aisheh, Razanne and Suneesh, Shyamli and Didiot-Cook, Tom and Jones, Simon and Sardinha, Emanuel Nunez and Munera, Marcela and Hauert, Sabine},
  booktitle={2025 34th IEEE International Conference on Robot and Human Interactive Communication (RO-MAN)},
  pages={2260--2265},
  year={2025},
  organization={IEEE}
}

@inproceedings{miyauchi2026human,
  title={Human Perceptions of Warmth and Competence in Swarm Robot Behavior},
  author={Miyauchi, Genki and Gro{\ss}, Roderich and Chen, Chaona},
  booktitle={Companion Proceedings of the 21st ACM/IEEE International Conference on Human-Robot Interaction (HRI), Late Breaking Reports},
  year={2026},
}

\end{document}